\documentclass{article}

\usepackage{PRIMEarxiv}

\usepackage[utf8]{inputenc} 
\usepackage[T1]{fontenc}    
\usepackage{hyperref}       
\usepackage{url}            
\usepackage{booktabs}       
\usepackage{amsfonts}       
\usepackage{nicefrac}       
\usepackage{microtype}      
\usepackage{lipsum}
\usepackage{fancyhdr}       
\usepackage{graphicx}       
\usepackage{amsmath}
\usepackage{subfigure}
\graphicspath{{media/}}     

\title{Bio-realistic Neural Network Implementation on Loihi 2 with Izhikevich Neurons}

\author{
  Recep Buğra Uludağ\\
  Özyeğin University \\
  Istanbul, Turkey \\
  \texttt{bugra.uludag@ozu.edu.tr} \\
   \And
  Serhat Çağdaş \\
  Yalova University \\
  Yalova, Turkey\\
  \texttt{serhat.cagdas@yalova.edu.tr} \\
  \And
  Yavuz Selim İşler \\
  Osmaniye University \\
  Osmaniye, Turkey\\
  \texttt{yavuzselimisler@osmaniye.edu.tr} \\
  \And
  Neslihan Serap Şengör \\
  İstanbul Technical University \\
  Istanbul, Turkey\\
  \texttt{sengorn@itu.edu.tr} \\
  \And
  Ismail Akturk \\
  Özyeğin University \\
  Istanbul, Turkey\\
  \texttt{ismail.akturk@ozyegin.edu.tr} \\
}

\begin{document}
\maketitle

\begin{abstract}
In this paper, we presented a bio-realistic basal ganglia neural network and its integration into Intel’s Loihi neuromorphic processor to perform simple Go/No-Go task. To incorporate more bio-realistic and diverse set of neuron dynamics, we used Izhikevich neuron model, implemented as microcode, instead of Leaky-Integrate and Fire (LIF) neuron model that has built-in support on Loihi. This work aims to demonstrate the feasibility of implementing computationally efficient custom neuron models on Loihi for building spiking neural networks (SNNs) that features these custom neurons to realize bio-realistic neural networks. 
\end{abstract}


\section{Introduction}

Spiking Neural Networks (SNNs) enable a new model of computing that mimics asynchronous and sparse spiking behavior of biological neurons. These spiking neurons maintain their own internal states stored in local memory. Their dynamics change over time based on their internal states and external stimuli they receive. Inspired by biological counterparts, spiking neurons try to build highly efficient, parallel and scalable computing substrate by realizing SNNs in a dedicated neuromorphic hardware, such as Intel's Loihi~\cite{davies2018loihi}, IBM's TrueNorth~\cite{akopyan2015}, SpiNNaker~\cite{painkras2013}, BrainScaleS~\cite{schemmel2010}, NeuroGrid~\cite{benjamin2014}, BrainDrop~\cite{neckar2019} and DYNAPs~\cite{moradi2018}. 
Each of these neuromorphic chip has its own benefits and limitations. However, all neuromorphic chips are inspired by the structural and functional features of the biological brain, aiming energy-efficient, fault tolerant, highly parallel and scalable computing substrate.

Programming neuromorphic chips and developing SNNs remain open, especially for applications that target bio-realistic behavior of neural networks. 
Therefore, in this paper, we demonstrate a showcase for developing bio-realistic neural network implementation on neuromorphic chip, Loihi in particular.  While TrueNorth has a similar hardware and programming support, its inter-neuron connectivity capability is relatively limited as opposed to Loihi which supports larger-scale connectivity. SpiNNaker has similar capabilities, but is constructed of standard CPU hardware. Loihi’s capabilities on the other hand are built-in on a chip which allows to explore new programming paradigms~\cite{dey2022}. 

For the showcase that we demonstrated in this paper, we picked basal ganglia (BG) as neural network since it is a network of interest of many researchers due its central role on computationally demanding tasks, such as decision-making and reward related learning. In particular, we developed an SNN to implement BG network that performs Go/No-Go task.  While Loihi provides built-in Leaky-Integrate-and-Fire (LIF) neuron model, we employ custom-built Izhikevich neurons (via microcodes that Loihi allows execute)  on this BG network to further demonstrate how custom neuron models can be implemented and used on Loihi.

\textbf{Related Work:} ~\cite{dey2022} demonstrates an SNN model for a mouse primary visual cortex using Leaky-Integrate and Fire (LIF) neuron on Loihi. In ~\cite{polykretis2020}, an SNN-based neuronal-astrocytic central pattern generator was implemented on Loihi to control the locomotion of a hexapod robot. In this work, bursting neuron dynamics were implemented on Loihi by using customizable multi-compartmental neurons, in particular using non-spiking compartments that had no voltage reset (these non-spiking compartments were essential for realizing a bursting neuron behavior)~\cite{polykretis2020}. On the other hand, various SNNs were developed to solve a diverse range of problems from various domains, such as event-based data processing, adaptive control, constrained optimization, sparse feature regression, and graph search~\cite{davies2021}. Among those, noticeable ones are inspired by bio-realistic neural networks responsible for brain's navigational system~\cite{tang2018}~\cite{tang2019}, 
and a canonical columnar cortical circuit which is considered to have impact on capabilities such as perceptual organization, motion detection and attention~\cite{lohr2020}.
Including these earlier studies, there are few realization of SNNs to implement bio-realistic neural networks on neuromorphic chips. Our hope is that this paper would demonstrate the feasibility of implementing bio-realistic neural networks on Loihi, helping to accelerate the advancement in the field.



We make the following contributions in this paper:
\begin{itemize}
    \item To achieve the bio-realistic functionality of neuron behavior, we employed the Izhikevich neuron model which has been implemented on Intel's Loihi using microcodes.
    \item We developed an SNN that features Izhikevich neurons to implement bio-realistic basal ganglia network running on low-power Intel's Loihi.
    \item We used the basal ganglia network to demonstrate a fundamental neuropsychological decision-making task, namely the Go/No-Go task, running on Intel's Loihi. 
    \item We quantify the performance and energy usage of Intel's Loihi for the implemented basal ganglia neural network, performing the Go/No-Go task.
\end{itemize}

We hope that this paper would demonstrate the feasibility and flexibility of using microcodes on Loihi neuromorphic chip to support different neuron types and mapping bio-realistic neural networks as SNNs. By doing so, we hope that this study would catalyze the participation of wider research community into neuromorphic computing.

\begin{figure*}[ht]
  \centering
  \includegraphics[width=0.75\linewidth]{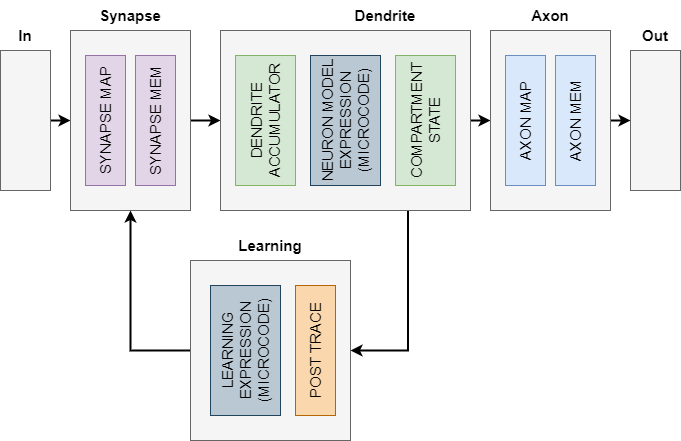}
  \caption{Simplified Diagram of Loihi 2 Neurocore Architecture.
}
  \label{fig:loihidiag}
\end{figure*}

This paper is organized as follows. We highlight the features of the Loihi architecture in Section~\ref{sec:loihi}. Section~\ref{sec:izhikevich-modelling} explains the Izhikevich neuron model and its implementation. We discuss the neural structure of basal ganglia along with its functionality and Loihi implementation in Section~\ref{sec:basal-ganglia}. Our experimental methodology and evaluation are provided in Section~\ref{sec:evaluation}. Finally, we conclude the paper with Section~\ref{sec:conclusion}. 


\section{Intel Loihi Architecture}
\label{sec:loihi}

Intel Loihi chip is a low-power, digital, many-core neuromorphic processor that was developed to mimic the complex functions of the human brain. It consists of a mesh of neuromorphic cores (a.k.a. neurocore), with x86 processing cores, and it features off-chip communication interface that allows to scale out to many other chips. 
As the building blocks of Loihi, neurocores can perform real-time inference and online learning (e.g., spike-timing-dependent plasticity (STDP) and reinforcement learning). 
These neurocores communicate asynchronously via barrier synchronization messages~\cite{davies2018loihi}. 

The Loihi architecture incorporates hierarchical connectivity, dendritic compartments, synaptic delays, and synaptic learning rules, in order to facilitate the implementation of SNNs in silicon. Each neurocore has synapse, dendrite, axon and learning components, as illustrated in Figure~\ref{fig:loihidiag}. The synapse block processes all the incoming spikes from the previous compartment/neuron and captures the synaptic weight from the memory. 
In the dendrite block, dendrite accumulators provide accumulated synaptic inputs. Compartment state keeps neuron-specific state that can vary over time and accumulated inputs are used to update these dynamic state variables.
The axon block generates the spike messages to be carried ahead by the fan out cores. The learning block updates the synaptic weights based on a learning rule. In this paper, we have not incorporated any learning rule and left it as a future work.

The learning block of Loihi provides standard STDP and a reward-based learning mechanism for online training. Its successor Loihi 2 supports three factor learning rule which allows to map the modulatory factors to post-synaptic neurons individually.  It also supports custom on-chip learning via a microcode-based learning rules that can be executed by each neurocore. To program these neurocores and implement SNNs, a programming framework called Lava~\cite{lava2021soft} can be used. Memory capacity of the blocks limits the total number of neurons that can be simulated by a neurocore. As the precision of the neuron model increases, the number of neuron per-neurocore decreases. 

First version of Loihi presents only generalized Leaky Integrate and Fire (LIF) neuron model. Despite its simplicity, LIF neurons may fall short to capture the dynamics of the bio-realistic neural networks which necessitate more sophisticated, yet computational efficient neuron models~\cite{beeler2010tonic}~\cite{gerstner2002}~\cite{schultz_07}. Because of that Loihi 2, successor of the first version, supports fully programmable neuron models that can be implemented via microcode instructions. The microcode instruction set includes bitwise, math and conditional operations. One can identify variables, parameters and define the behavior of the neuron model by using short sequence of instructions~\cite{orchard2021efficient}. Because of the same drawback of LIF models, we preferred to use Izhikevich neuron model in this work \cite{izhikevich2003simple}. We exploit these features of Loihi 2 in implementing Izhikevich neurons and furthermore the basal ganglia neural network.

\section{Implementation of Neuron Model}
\label{sec:izhikevich-modelling}

\begin{figure}[ht]
  \centering
  \includegraphics[width=0.7\linewidth]{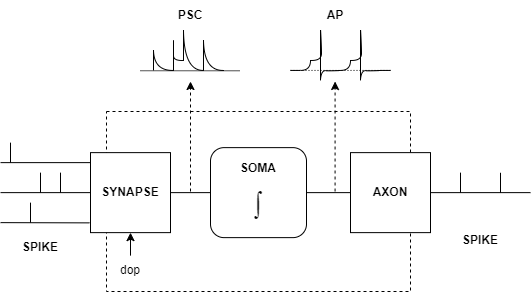} 
  \caption{Single point neuron diagram of Izhikevich neuron model given in Eq. \ref{eq:IZH_EULER} and Eq. \ref{eq:i-sync}. Post-synaptic current (PSC) is computed using spike inputs. PSC causes postsyanptic potential and action potential (AP) at soma calculated using Eq. \ref{eq:IZH_EULER}. Finally, action potentials are converted to spikes and relayed to other neurons via axon.}
  \label{fig:mapping-neuron-inputs}
\end{figure}

The following equations describe the discrete Euler form of the Izhikevich neuron model which is the basis of the neuron model used in this paper for Loihi implementation:

\begin{equation}
\begin{aligned}
v[t +\Delta t]  &= v[t] + (0.04v[t]^2+5v[t]-u[t]+I) \Delta t\\
u[t +\Delta t]  &= u[t] + a(bv[t] - u) \Delta t \\ 
v[t]>v_{p} &\Rightarrow  v[t+\Delta t] = c, u[t+\Delta t] = u[t] + d
\end{aligned} \label{eq:IZH_EULER}
\end{equation}

where $v$ is the membrane potential and $u$ is recovery variable, 
 $v_p$ is peak potential value corresponding to $v_{thr}$ as in~\cite{izhikevich2006} and besides the initial values of $v$, the value of $v_p$ are the only biologically defined parameters. Behavior of a neuron can be tuned by setting the parameters of $a$, $b$, $c$ and $d$. $I$ is the current that consists of an external component and synaptic input, which can be described as:

\begin{equation}
\begin{aligned}
I[t]  &= I_{const}[t] + I_{syn}[t] (1 + \beta \Delta dop ) \\
I_{syn}[t + \Delta t] &= \alpha I_{syn}[t] +   \sum{(w_{ij} \delta(t-t^{(i)}_f))}
\end{aligned}
\label{eq:i-sync}
\end{equation}

where $t^{(i)}_f$ is the spike times of $i$th pre-synaptic neuron and $w_{ij}$ is the strength of this synapse, whereas $I_{syn}$ is used as a dynamic quantity to define synaptic currents.
When a pre-synaptic neuron fires, the current increases momentarily by the value of synaptic weight. In the equation, decay parameter $\alpha = 1 - \Delta t / \tau$ is a function of exponential time constant $\tau$ ($15ms$ in our case) and timestep duration $\Delta t$ (which is $1/8ms$). Finally, the dopamine modulation is achieved by a factor called $\beta$~\cite{chersi2013spiking} which regulates synaptic transmission between pre- and post-synaptic neurons. If $\beta$ is higher than the baseline value, it increases the excitability of post-synaptic neurons. On the other hand, if the $\beta$ value is negative, then it has an opposite effect on the post-synaptic neurons. Figure~\ref{fig:mapping-neuron-inputs} illustrates the relationship among blocks and how inputs to the neuron model is processed and how output spike is generated, following the formulation provided above.

Computation of the neuron model is expressed in dendrite block of the neurocore (Figure~\ref{fig:loihidiag}). Variables and neuron specific parameters in the model are stored in the compartment states. Each compartment state has small amount of memory. As the complexity of the model or the precision of the computation increases more memory is needed. Thus, multiple states are occupied. In this study, all the state variables $(v,u, Iext, Isyn)$ of the model are stored in the compartment state. In addition, $a,b,c$ and $d$ parameters are also kept in the compartment state to be able to provide variability between neurons in the group (Table \ref{tab:CompState}).

\begin{table}
  \caption{Memory allocation in compartment state.}
  \label{tab:CompState}
  \centering
  \begin{tabular}{lcc}
    \toprule
     &Size (Bits)&Compartment State Index\\
    \midrule
    v & 24& 2\\
    u & 24& 2\\
    a & 16& 0\\
    b & 16& 0\\
    c & 24& 1\\
    d & 24& 1\\
    Isyn & 24& 0\\
    Iconst & 16& 2\\
    delta\_dop & 16 & 1 \\ 
  \bottomrule
\end{tabular}
\end{table}

\begin{figure*}[ht]
  \centering
\subfigure{\includegraphics[width=\linewidth]{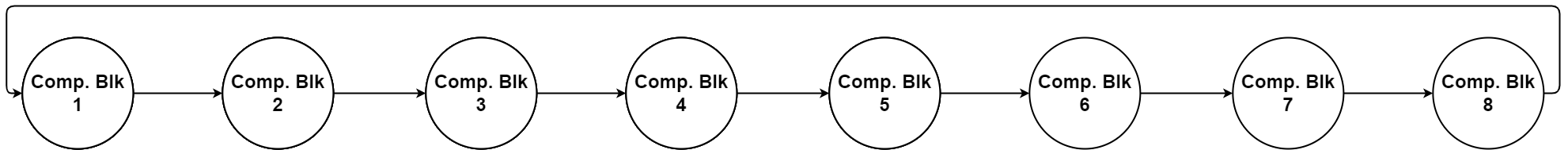}}
(a)
\vspace{5mm} 
\subfigure{\includegraphics[width=\linewidth]{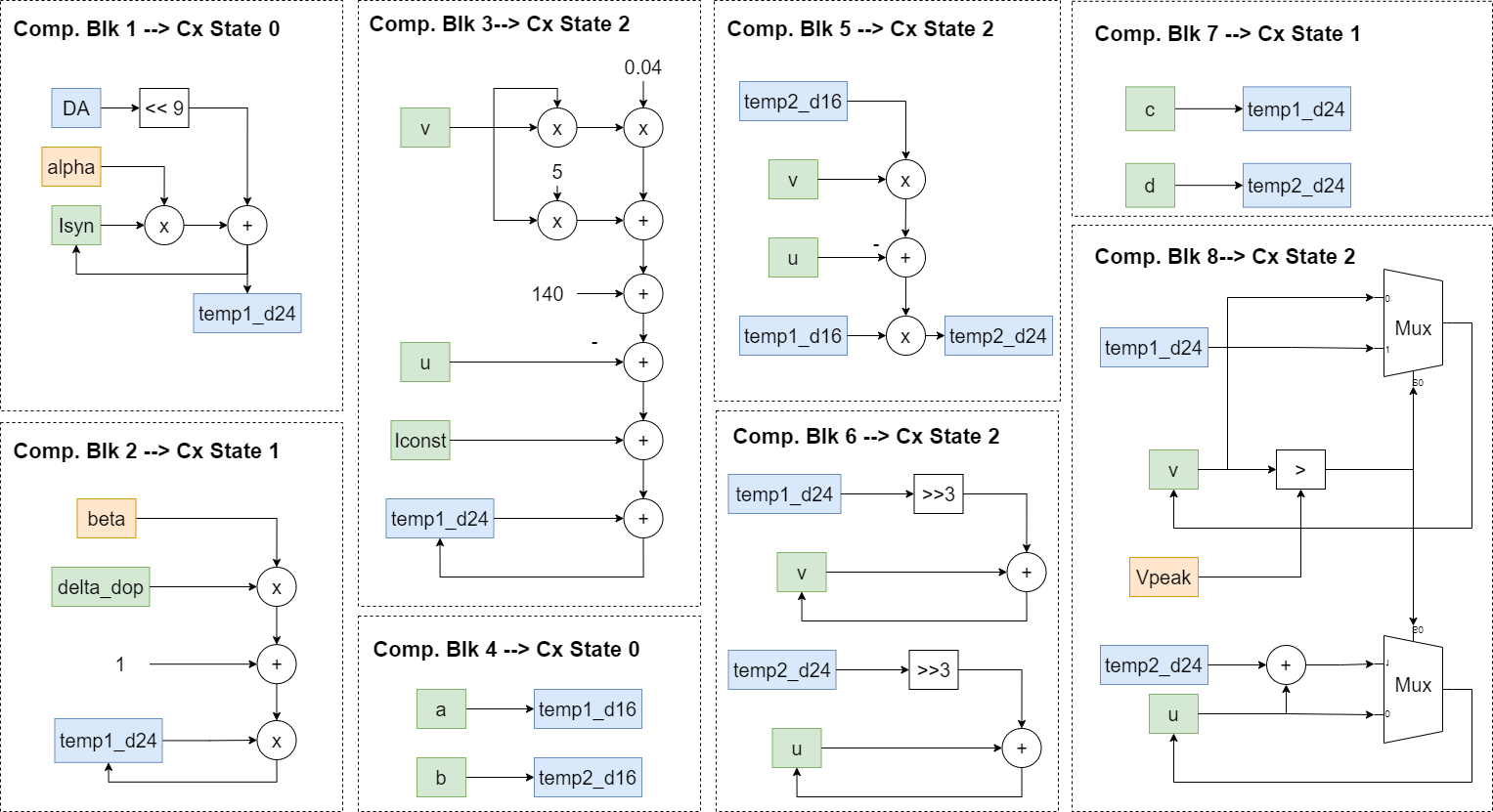}}
(b)

  \caption{ (a) Overall view of block diagram of Izhikevich neuron discrete computation for Loihi 2;  (b) each block executes limited number of instructions and is related to just one compartmental state. Variables with green color allocate compartment state and blue colored blocks represent allocation of register memory. Orange variables are constant. The same constant value is used by all cells in a neuron group.}
  \label{fig:IzhLoihi}
\end{figure*}

Figure~\ref{fig:IzhLoihi}, shows the block diagram of model's discrete computation on the hardware. Instructions are executed in discrete computation blocks. In each block one compartment state is accessible. Data of the state is temporarily stored in registers when needed to overcome this constraint. Some blocks (such as Blk 4, Blk 7) are used just for that purpose. Total synaptic input activity for the block diagram is provided through dendrite accumulator block which is denoted by DA in Figure~\ref{fig:IzhLoihi}. 12 bits of the data is used as fraction for all variables except DA. Accumulated value is shifted nine bits initially since its fraction bit number is three. In Blk 6, data is shifted three bits instead of multiplying by $\Delta t = 1/8$ before updating state variables. 

\begin{figure*}[ht]
  \centering
 \subfigure{\includegraphics[width=0.33\linewidth]{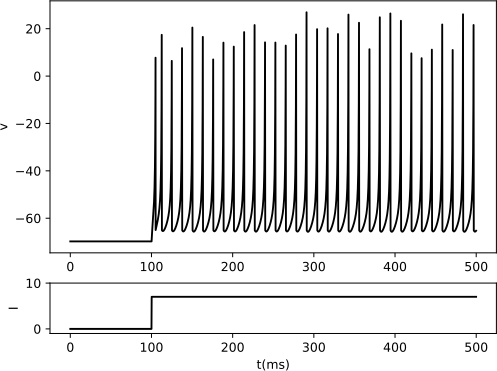}}
 \subfigure{\includegraphics[width=0.33\linewidth]{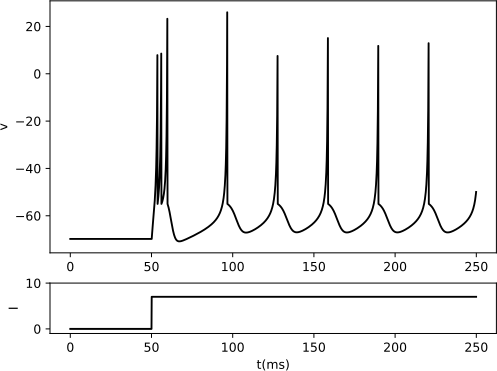}}
 \subfigure{\includegraphics[width=0.33\linewidth]{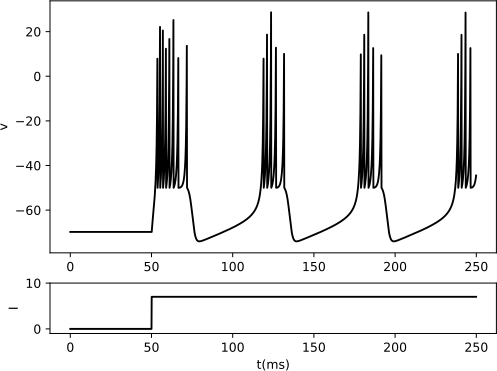}}
 \subfigure{\includegraphics[width=0.33\linewidth]{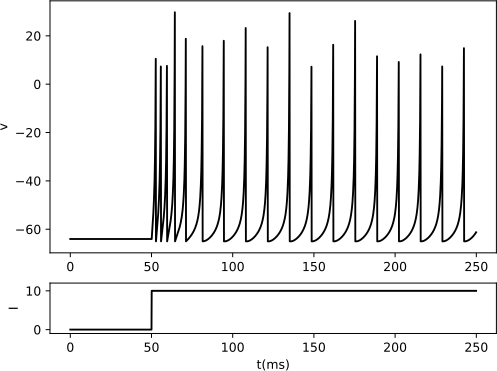}}
 \subfigure{\includegraphics[width=0.33\linewidth]{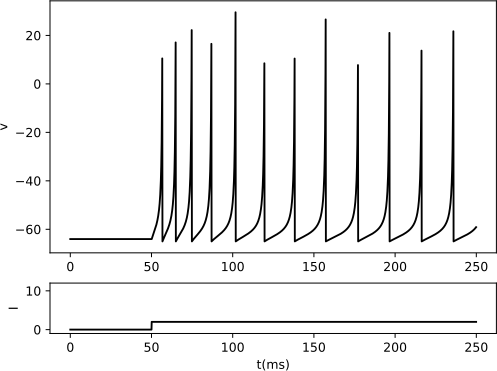}}
 \subfigure{\includegraphics[width=0.33\linewidth]{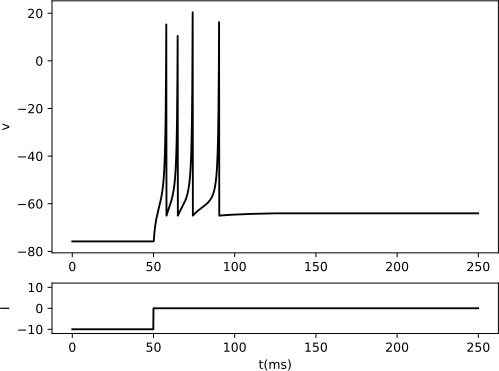}}
 
  
  \caption{Different behavior mimicked by Izhikevich neuron model simulated on Loihi2 neurocore. Top: fast spiking, intrinsically bursting, chattering. Bottom : low-threshold spiking neuron, tonic and rebound spiking thalamo-cortical cell    }
  \label{fig:SpRegime}
\end{figure*}

After the implementation, Izhikevich neurons are tested to check their ability of mimicking different spiking regimes. As shown in Figure~\ref{fig:SpRegime}, the neuron simulated on Loihi 2 is able to show different spiking behavior like regular spiking, fast spiking, intrinsically bursting, chattering, low-threshold spiking, tonic and rebound spiking successfully.

\begin{figure*}[h]
  \centering
  \subfigure{\includegraphics[width=0.45\linewidth]{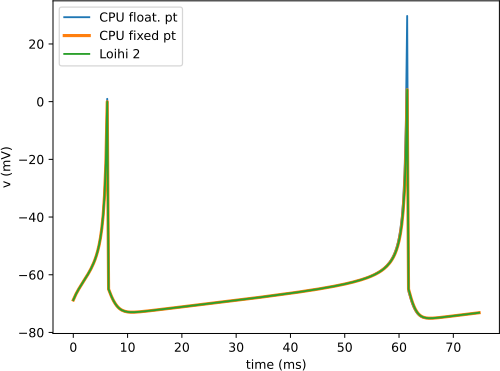}}
 \subfigure{\includegraphics[width=0.45\linewidth]{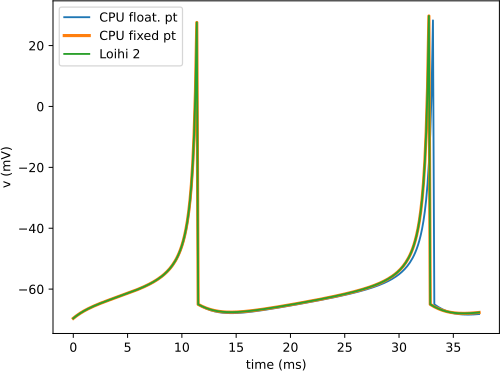}}
  \caption{Spike timing comparison of neurons with regular spiking (left)  and fast spiking (right) behavior for CPU simulation using floating point, fixed point and Loihi 2 implementation.}
  \label{fig:RSFS}
\end{figure*}

In addition, the behavior of Izhikevich neuron model running on Loihi 2 chip is compared to the behavior of the same model simulated on CPU using both floating point and fixed point numbers. For the accuracy analysis, the ERRt measure is used which is based on spiking times. ERRt is defined as the time interval differences between two spikes in the original and the proposed model~\cite{heidarpour2016cordic, heidarpur2019cordic}.
The first spikes in the two simulations are synchronized for the ERRt calculation. Afterwards, the ratio of the difference between the second spikes to the difference between the two spikes of a neuron is used as a measure of accuracy:

\begin{equation}
\begin{aligned}
ERRt  &= \left|  \frac{\Delta t_x - \Delta t_{fl}}{\Delta t_{fl}}  \right|\times 100\\
\Delta t_x  &= t^{(x)}_{s2} - t^{(x)}_{s1} \\  
\end{aligned} \label{eq:ERRT}
\end{equation}

In Eq.~\eqref{eq:ERRT}, $t^{(x)}_{s1}$ and $t^{(x)}_{s2}$ are the time of first and second spikes, respectively. $\Delta t_{x}$, on the other hand, is the difference of these spikes. As seen in the equation, simulation result with floating point (i.e., $\Delta t_{fl}$) is used as the reference. Two different analysis showing regular spiking (RS) and fast spikig (FS) behavior are carried out for each implementation as shown in Figure~\ref{fig:RSFS}. On the other hand, Table~\ref{tab:ERRT} shows the ERRt values for these analysis. For RS behavior, all three implementations spike at the same time whereas FS spiking neurons differ in floating point implementation. In both RS and FS analysis, neurons implemented with fixed point on CPU and Loihi 2 yield consistent results (which increases the confidence on the fidelity of neuron behavior observed on Loihi 2).

\begin{table}
  \caption{Accuracy Analysis (ERRt) for CPU Fixed Point Simulation and LOIHI 2 Hardware Implementation}
  \label{tab:ERRT}
  \centering
  \begin{tabular}{lcc}
    \toprule
     &RS&FS\\
    \midrule
    CPU float. pt. - CPU fixed pt.& $\%0$ & $\%1.724$ \\ 
CPU float. pt. - Loihi 2 & $\%0$ & $\%1.724$ \\
  \bottomrule
\end{tabular}
\end{table}

Loihi employs a fixed-size discrete time-step to simulate the dynamics of the models. In this study, we use an explicit Euler integration scheme, where the time steps relate to the algorithmic time of the computation. Therefore, the algorithmic time may differ from the execution time spent on the hardware~\cite{dey2022}. Moreover, due to hardware constraints and to increase the efficiency of the Loihi chip, certain bit-size constraints are imposed on the state variables which should be taken into account during microcode development to preserve the fidelity of the model that captures the intended dynamics of the neurons.

\section{Basal Ganglia Circuit and Go/No-Go Task}
\label{sec:basal-ganglia}

\begin{figure}[ht]
  \centering
  \includegraphics[width=0.6\linewidth]{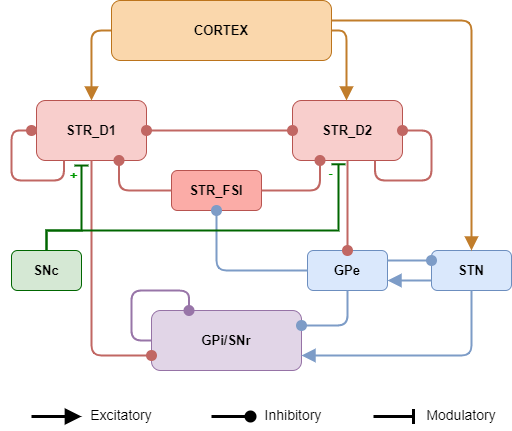}
  \caption{The diagram of the cortico-basal ganglia circuit. STR-D1: direct striatum pathway neuron, STR-D2: indirect striatum pathway neuron, STR-FSI: striatum fast spiking neuron, STN: subthalamic nucleus, GPe: globus pallidus external,SNc: subtantia nigra pars compacta, SNr/GPi: substantia nigra pars reticulata/globus pallidus internal.}
  \label{fig:bg-pathways}
\end{figure}

Basal Ganglia (BG) is a collection of nuclei stationed between the top of midbrain and the base of forebrain. It takes many afferent connections from other regions of the brain but primarily the cortex. There are unique parallel loops that encompasses basal ganglia identified depending on their source and target, which result in various neuropsychological functions such as motor control and learning, occular movement, limbic and cognitive behaviours~\cite{alexander1990}. While each loop goes through BG input nuclei in different regions and project to different output regions, the underlying algorithm and structure remains similar. Figure~\ref{fig:bg-pathways} illustrates the main pathways of cortico-basal ganglia network and indicates excitatory, inhibitory and modulatory connections of the BG network. 

The basal ganglia has two pathways, the direct and indirect, which emerge from the main input nuclei, striatum. The direct pathway facilitates movement by sending a signal to the output nuclei while the indirect pathway suppresses movement by sending a signal to the inhibitory pathway. In addition, the basal ganglia has a hyperdirect pathway that bypasses the striatum and directly inhibits movement, which is involved in pre-movement inhibition. By using inner connection between these multiple pathways, the lateral inhibition of losing actions over winning action is realized. Thus, the interplay of two pathways, namely, direct and indirect pathway will be considered in this paper to realize a computational model of \textit{Go/No-Go} task. This task is used as neuropsychological test to measure the ability of inhibiting habitual responses in favor of novel one.


The classical model of BG introduces many neuroanatomical and functional diversity both in terms of the neuron behaviour, neurotransmitters and circuitry. While each connection provides unique interactions, the task can be implemented using a simplified model, relying on Izhikevich neurons. 

The details of neuron parameters and their connections to implement BG model on Loihi are given in Table~\ref{tab:bgpar} and Table~\ref{tab:bgcon} (mostly derived from  ~\cite{sen2018building}), respectively.  Each population is limited to 100 neurons, as in~\cite{goenner2021spiking}, to keep the mapping time of the network onto neurocores reasonable. $I_{const}$, is added to each of the neurons to increase the basal activity~\cite{sen2018building}\cite{thibeault2013}.

The afferent connections from the cortex is simulated through a Poisson spike generator that is microcoded via utilizing the random registers provided by the neurocore, whose parameters are shown in Table~\ref{tab:cortexcon}. Since SNc is not explicitly implemented, the impact of dopamine is represented as a variable in the neuron model. 

\begin{table}
  \caption{Neuron Parameters For the Basal Ganglia Circuit}
  \label{tab:bgpar}
    \centering
  \begin{tabular}{lcccccc}
    \toprule
    Group	& $a$&$b$&$c$&$d$&$I_{const}$&$\beta$\\
    \midrule
STR\_D1&0.02 &0.2 &-65 &8&-&0.6 \\  
STR\_D2&0.02 &0.2 &-65 &8&-&-0.6\\  
STR\_FSI&0.1 &0.2 &-65 &2&-&0 \\
GPe&0.1 &0.585 &-65 &4& 5&0  \\  
STN&0.005&	0.265&	-65& 2&2&0 \\  
GPi/SNr&0.005&	0.32&-65&2&5&0\\  
     
  \bottomrule
\end{tabular}
\end{table}

\begin{table}
  \caption{Connectivity Parameters of Basal Ganglia Circuit }
  \label{tab:bgcon}
    \centering
  \begin{tabular}{cccc}
    \toprule
    Presyn. Group & Postsyn. Group & Weight & Conn. Prob.\\
    \midrule
    STR1 & STR1 & -0.3 & 1\\
         & STR2 & -0.3 & 1\\
         & GPi/SNr  & -7.5 & 0.15\\
    STR2 & STR1 & -0.3 & 1\\
         & STR2 & -0.3 & 1\\
         & GPe  & -7.5 & 0.15\\
    STR\_FSI & STR1 & -1.5 & 0.1\\
        & STR2 & -1.5 & 0.1\\ 
    GPe & STR\_FSI & -2.25 & 0.1\\
        & GPe & -2.25 & 0.1\\
        & STN & -2.25 & 0.1\\
        & GPi/SNr & -2.25 & 0.1\\
    STN & GPe & 2.25 & 0.1\\
        & GPi/SNr & 2.25 & 0.1\\
    GPi/SNr & GPi/SNr & -1 & 0.1\\
     
  \bottomrule
\end{tabular}
\end{table}

\begin{table}
  \caption{Poisson Spike Generator Groups and Connection Parameters }
  \label{tab:cortexcon}
    \centering
  \begin{tabular}{ccccccc}
    \toprule
    Presyn. & Postsyn. & Weight & Conn. & Freq. (Hz) \\
    Group &  Group &  & Prob. & \\
    \midrule
    G\_Ctx1  & STR1 & 5    & 0.2&15\\
    G\_Ctx2  & STR2 & 5    & 0.2&15\\
    G\_Ctx3  & STN & 1.125 & 0.05&4\\
    G\_noise & All & 0.05  & 0.1&5\\ 
  \bottomrule
\end{tabular}
\end{table}

\section{Evaluation}
\label{sec:evaluation}

\begin{figure*}[ht]
\centering  
\subfigure{\includegraphics[width = 0.49\linewidth]{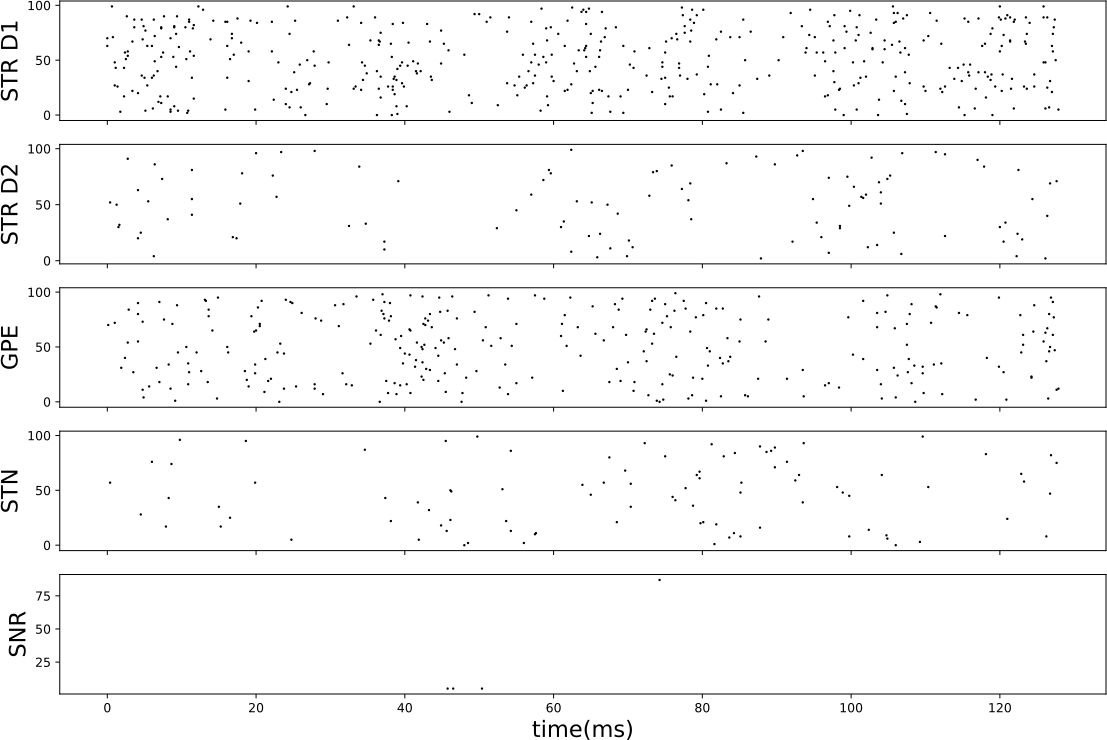}}
\subfigure{\includegraphics[width = 0.49\linewidth]{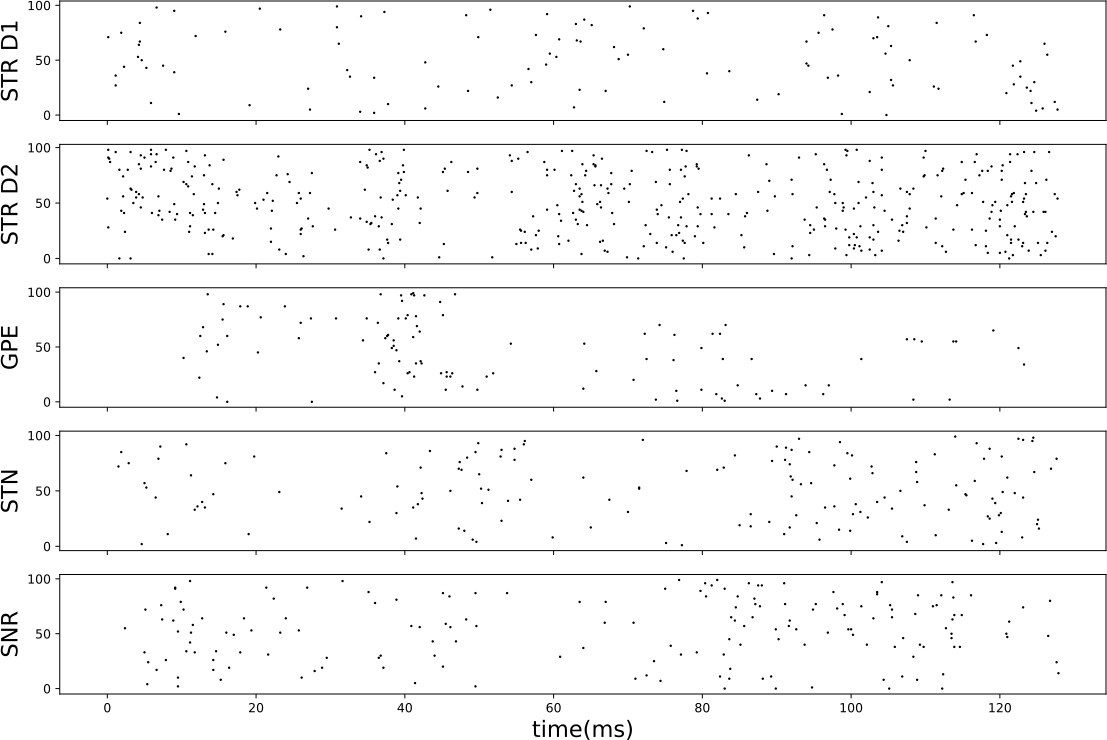}}
  \caption{Raster plots of striatum (MSN D1 and D2), GPe, STN and GPi/SNr neuron groups when (left) the dopamine release is higher and (right) the dopamine release is lower than the baseline level.}
  \label{fig:raster-plot}
\end{figure*}

\begin{table*}[t]
  \caption{Performance Analysis of Various Networks }
  \label{table:performance}
  \begin{tabular}{cccccccc}
    \toprule
     &  &  &  & Latency & Energy & Energy per & \\
     & Power  & Power  & Power   & per   & per  & Inference &   Energy Delay  \\
     Paper/Model &   Static     & Dynamic        &  Total      & Inference  &  Inference  & Dynamic & Product  \\
    & (W) &  (W) & (W) & ($\mu$s) & ($\mu$J) &  ($\mu$J) & ($\mu$Js) \\
    \midrule
    BG Base Level Dopamine  & 0.537 &	0.045 &	0.582 & 7.232	& 4.206 & 0.322 &	3e-5\\
    BG High Dopamine &  0.536 & 0.049 & 0.585 & 7.232  & 4.232 & 0.356 & 3.1e-5\\
    BG Low Dopamine & 0.537 &	0.045 & 0.582 & 7.228 &	4.210 & 0.328 &	3e-5\\
    SpiNNaker \cite{sen2018building} 1CH & n/a &	n/a & 0.8 & 1e3  & 80 & n/a &	0.08\\
    SpiNNaker \cite{sen2018building} 3CH & n/a &	n/a & 1.8 & 1e3  & 180 & n/a &	0.18\\
    SpineML \cite{sen2018building} 3CH & n/a &	n/a & 95 & 2.67e3  & 2.53e5 & n/a &	6.77e2\\
  \bottomrule
\end{tabular}
\end{table*}

To facilitate the development of SNNs for capturing bio-realistic behavior of neural networks, a software programming platform, called Lava, has been introduced recently~\cite{lava2021soft}. Lava framework allows to develop SNNs that employ different neuron models and these SNNs can be mapped to different neuromorphic hardware backends (it is also possible to simulate the behavior of the SNNs on CPU and GPU). In our case, we use Lava to develop an SNN that implements Basal Gaglia (BG) network targeting Loihi as a hardware backend. 

In our analysis, the BG circuit was run in three configurations based on the level of dopamine is introduced: i) base-level dopamine, ii) low-level dopamine and iii) high-level dopamine. In these configurations, the input from cortex remains constant, however, different levels of dopamine is introduced to striatum neurons. The Figure~\ref{fig:raster-plot} illustrates the basal ganglia input, intrinsic and output nuclei activity in high dopamine and low dopamine stimulations. 
As it can be seen, high level dopamine excitation of D1R type MSNs of striatum results in decrease in the inhibitory activity of SNr that allows action selection. In the opposite, low level dopamine causes the promotion of indirect pathway and suppression of the direct pathway causing total suppression of the action. This demonstrates the realization of intended functionality of BG circuit for Go/No-Go task.

For quantitative analysis, we examined different metrics, including resource utilization, power dissipation, energy usage and execution time on Loihi 2 and compared the results with similar BG circuits running on other platforms. In particular, the same BG circuit is used in~\cite{sen2018building} that has both SpiNNaker neuromorphic chip and simulation-based CPU (referred as SpineML in text) implementations.
While the BG circuit is the same, the number of neurons in the  populations are smaller in our implementation. Moreover, they have three channels in their implementation where the replicated BG models are connected via SNr and GPe to apply Go/No-Go task over multiple actions. Table~\ref{table:performance} shows the comparison for different metrics, and their implementations with one and three channels are labeled as 1CH and 3CH, respectively. Our implementation has only one channel. 

All measurements were obtained using Lava version 0.4.1 on Oheo Gulch that incorporates a single socketed Loihi 2 chip. Table~\ref{table:performance} shows breakdown of performance metrics across BG models on Loihi2,  SpiNNaker and CPU \cite{sen2018building} (Core i7 2600, 3.4 GHz).

As shown in Table~\ref{table:performance}, the BG circuit running on Loihi 2 is faster compared to SpiNNaker and CPU implementations. In terms of power dissipation, both Loihi 2 and SpiNNaker are comparable, and they both perform better than CPU implementation. While SpiNNaker has comparable power dissipation, Loihi 2 has lower energy-delay product, since each inference takes significantly less amount of time on Loihi 2. 

The implementation occupies 13 neurocores on Loihi 2, with 1750 neurons and 42,400 synapses. For the base-level dopamine configuration, a single time step takes approximately 7.25 $\mu$s and consumes  4.23 $\mu$J of energy (0.3 $\mu$J of which is active energy), resulting in an energy delay product of  3.1e-05 $\mu$Js.

Despite its high energy-efficiency and performance, we encountered some challenges and limitations during the deployment of the BG network on Loihi 2 (mostly due to limitations imposed by the current version of the Lava framework, rather than Loihi hardware itself). First, during the deployment of SNNs onto Loihi 2 for larger network models that feature larger number of neurons, we observed relatively longer  mapping duration due to increased synaptic connections (approx. an hour). Another limitation is the realization of conductance-based synaptic dynamics on Loihi2. The current version of Lava (0.4.1) allows only one dendritic accumulator input per neuron. This prevents us to use different reverse potential values for excitatory and inhibitory neurons. Because of that, it is not possible to compute the dynamics of different receptors, such as NMDA, AMPA, GABA~\cite{chersi2013spiking}. To overcome this limitation in our BG model,  we introduce a statically determined value $I_{syn}$ (as used in Eq.~\ref{eq:i-sync}).

Also, because of the current limitations on Lava,  
we had to implement a custom monitoring tool via microcode to collect desired neuron states and variables. We allocated subset of neurocores to buffer spike data (where these data are collected via microcode). These buffers are readout in certain time intervals by the host machine to obtain raster plots (similar to Figure~\ref{fig:raster-plot}).  

Overall, Loihi 2 provides flexible and energy-efficient neuromorphic substrate to have bio-realistic implementations of neural networks, practical design and deployment on Loihi 2. As the Lava framework matures, it would address the existing limitations and make it more convenient to explore advanced features of Loihi 2.

\section{Conclusion}
\label{sec:conclusion}

In this paper, we demonstrated that that Loihi is quite efficient in terms of performance and energy usage in the context of large bio-realistic neural network implementation and flexible in terms of supporting different neuron models featuring rich repertoire of neural dynamics by facilitating microcodes. In particular, we demonstrate a showcase of implementing a simplified bio-realistic basal ganglia neural network that carries Go/No-Go task, by using Izhikevich neurons on Intel Loihi chip.


\bibliographystyle{unsrt}  
\bibliography{references}

\end{document}